\documentclass{article}
\usepackage{spconf,amsmath,graphicx, bm}

\usepackage{times}
\usepackage{epsfig}
\usepackage{amssymb}
\usepackage{caption}
\usepackage{subcaption}
\usepackage{xcolor}
\usepackage{pifont}
\usepackage{adjustbox}
\usepackage{setspace}
\usepackage{mathtools}

\usepackage{booktabs}
\newcommand*\rot{\rotatebox{90}}
\usepackage{array,multirow}
\usepackage{rotating}
\usepackage{makecell}
\usepackage{tabularx}

\title{Two-stream Decoder Feature Normality Estimating Network\\ for Industrial Anomaly Detection}
\name{Chaewon Park \quad
   Minhyeok Lee \quad
   Suhwan Cho \quad
   Donghyeong Kim \quad
   Sangyoun Lee$^{*}$ \quad}
\address{Yonsei University, Seoul, Korea
\\{\tt\small \{chaewon28, hydragon516, chosuhwan, 2donghyung87, syleee\}@yonsei.ac.kr} \vspace{-0.2cm}
}
\begin{document}
%
\maketitle
\begin{abstract}
Image reconstruction-based anomaly detection has recently been in the spotlight because of the difficulty of constructing anomaly datasets. These approaches work by learning to model normal features without seeing abnormal samples during training and then discriminating anomalies at test time based on the reconstructive errors. However, these models have limitations in reconstructing the abnormal samples due to their indiscriminate conveyance of features. Moreover, these approaches are not explicitly optimized for distinguishable anomalies. To address these problems, we propose a two-stream decoder network (TSDN), designed to learn both normal and abnormal features. Additionally, we propose a feature normality estimator (FNE) to eliminate abnormal features and prevent high-quality reconstruction of abnormal regions. Evaluation on a standard benchmark demonstrated performance better than state-of-the-art models. 
\end{abstract}

\vspace{-0.3cm}
\section{Introduction}
\label{Section1:Intro}
\vspace{-0.3cm}
Anomaly detection is a computer-vision task that discriminates whether an image contains anomalies that deviate from the normal appearance. Automated anomaly detection has been gaining attention recently due to the increase in automatized systems in various industries, especially in quality control, medical treatment, and surveillance. 

Due to the data imbalance, meaning that abnormal samples are hard to obtain due to its rare occurrence, anomaly detection is formulated as a one-class learning setting where public datasets consist of a training set containing only normal images and a testing set containing both normal and abnormal images~\cite{survey}. Thus, it is desirable to train anomaly detection to distinguish normal features in the anomaly-free normal training dataset and identify the abnormal data derived from the learned feature distribution during inference. This is known as the unsupervised method in anomaly detection, used by most recent studies~\cite{riad, mvtec, smai, itad, kim2022fapm}.

	\begin{figure}[!t]
		\begin{center}
			\includegraphics[width=0.8\linewidth]{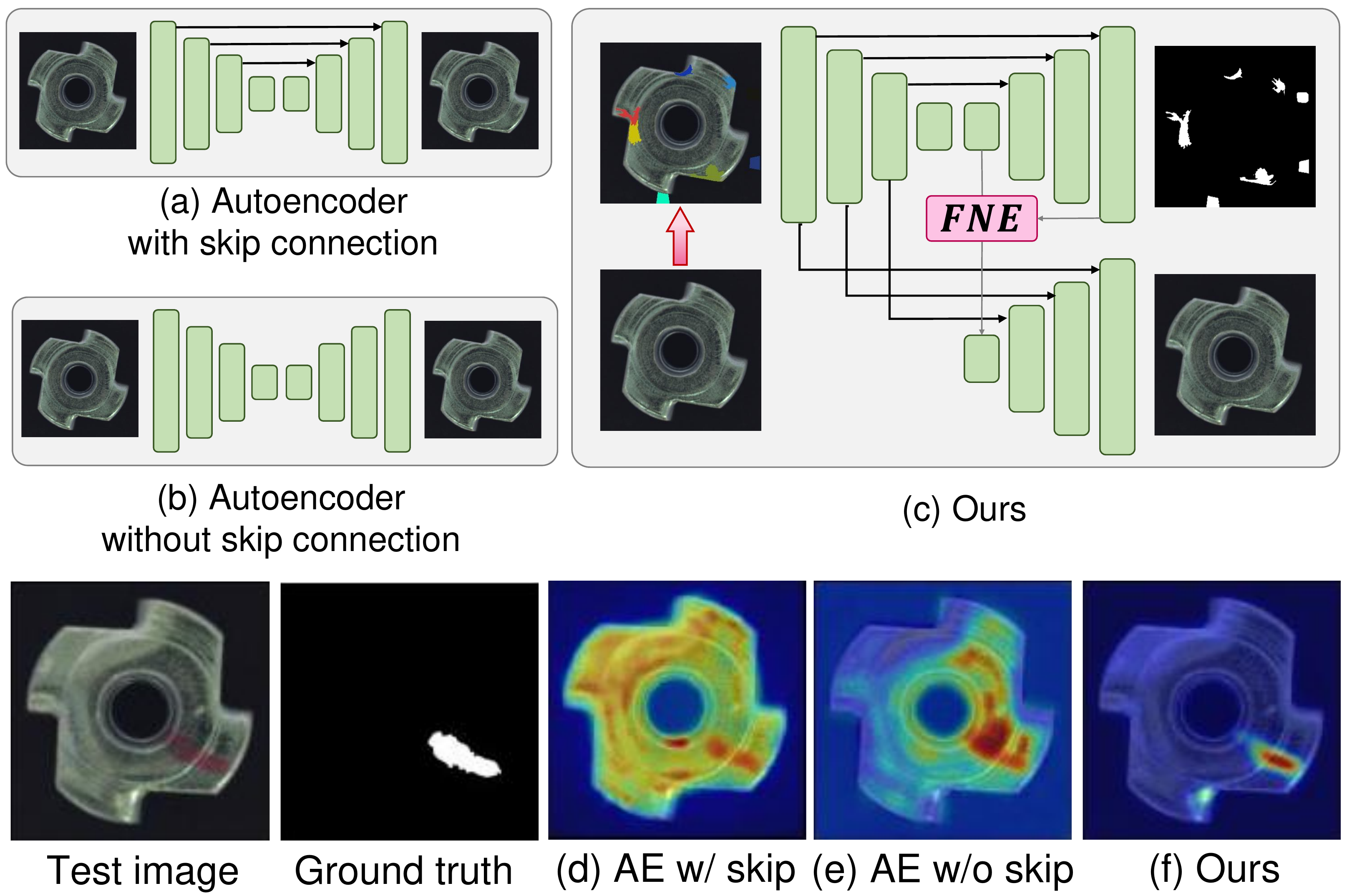}
		\end{center}
		\vspace{-0.5cm}
		\caption{Comparison of our model to the baseline models. (a) autoencoders with skip connections, (b) a general autoencoder, and (c) the proposed model. (d), (e), and (f) denote the respective predicted abnormal areas of each network.}
		\label{FIG1}
		\vspace{-0.7cm}
	\end{figure}

Many previous studies~\cite{mvtec, vevae, gan, itad, hou2021divide} relied on generative models to effectively reconstruct normal regions and fail on abnormal regions to discriminate the abnormal portions.  
These approaches have greatly improved anomaly detection performance. Specifically, \cite{itad} used an autoencoder (AE) similar to U-net~\cite{unet} to supplement the encoder features for better image reconstruction. However, this is problematic when the AE is over-fitted to the reconstructing task, leading to a perfect generalization of the abnormal test data too. 

	\begin{figure*}[!t]
		\begin{center}
			\includegraphics[width=0.85\linewidth]{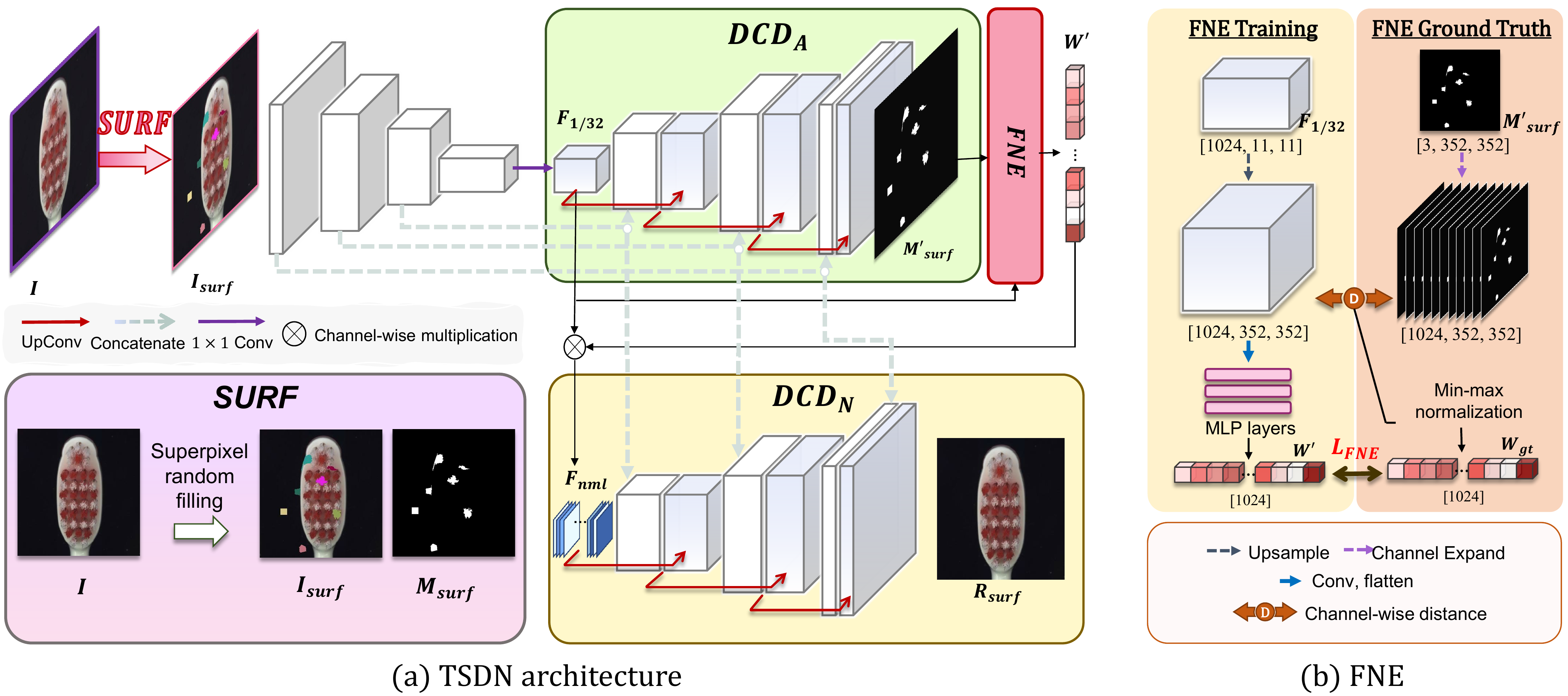}
		\end{center}
		\vspace{-0.5cm}
		\caption{(a) Model architecture. (b) The feature normality estimator (FNE). $DCD_A$ and $DCD_B$ in (a) learns the abnormal and normal features of the input image, respectively. (b) eliminates the abnormal features within the encoder feature block.}
		\label{FIG2:model}
	\vspace{-0.6cm}
	\end{figure*}

In Fig.~\ref{FIG1}, we demonstrate the problem of using AEs by showing two types of AEs, as shown in Figs.~\ref{FIG1} (a) and (b), and our model, shown in Fig.~\ref{FIG1} (c), as well as their anomaly localization results in Figs.~\ref{FIG1} (d), (e), and (f), respectively. 
The figure shows that the general AE itself in Figs.~\ref{FIG1} (b) and (e) is weak in anomaly detection. This deficiency is because excessive training of AEs leads to excellent reconstruction of the abnormal images due to the strong generalization capacity of convolutional neural networks. When skip connections are added, as in the U-Net~\cite{unet} architecture in Fig.~\ref{FIG1} (a), the performance degrades despite the additional details given to the decoder, as shown in Fig.~\ref{FIG1} (e). This degradation results from features of the abnormal regions unnecessarily conveyed to the decoder along with the features of the normal regions. Furthermore, these AEs do not act appropriately when abnormal images are fed during testing because they have never been trained on how to act for these images.

Therefore, we propose a two-stream decoder network (TSDN) that maximizes the advantages of skip connections in the U-Net~\cite{unet} architecture--the conveying of features lost during downscaling--and minimizes the disadvantages they have in anomaly detection. Fig.~\ref{FIG1} (c), shows the simplified architecture of our model, and the anomaly detection results are shown in Fig.~\ref{FIG1} (f). We propose a superpixel-based transformation method that we call superpixel random filling (SURF) to generate fake anomalies within the training data. This is intended to optimize the TSDN for anomalies. Moreover, the proposed two-stream decoder architecture works to learn both normality and abnormality by predicting the abnormal regions where SURF is applied and reconstructing the original anomaly-free image. 
Furthermore, we propose a feature normality estimator (FNE) module attached between the two decoders to alleviate conveying the unnecessary features of the abnormal regions. It recognizes the abnormal features and eliminates them to generate a refined normality-only feature map by suppressing the abnormal channels of the feature map. Thereby, we convey the details of the normal areas to the decoder by the skip connections and remove the useless features of the abnormal portions by the FNE. 
Consequently, we obtain a clear reconstruction of the original image.


\vspace{-0.5cm}
\section{Proposed Method}
\vspace{-0.3cm}
\subsection{Overview}
\vspace{-0.3cm}
The framework of TSDN is shown in Fig.~\ref{FIG2:model} (a).
Before feeding the input into the network, we distort the image by the proposed SURF, by which we obtain an abnormal image from the anomaly-free training data. Two decoders, namely $DCD_A$ and $DCD_N$, are proposed to learn the abnormality and normality of the input image, respectively. These two decoders are connected by a module called the FNE. During inference, SURF is detached, and only the $DCD_N$ output is used to estimate the normality of the input.

\vspace{-0.3cm}
\subsection{Superpixel Random Filling}
\vspace{-0.2cm}
We propose SURF to generate fake anomalies within the normal training data. This promotes a training phase similar to the testing scheme where both normal and abnormal images are input. 
To effectively mimic the common abnormal patterns, such as contamination, scratches, and cracks, we first split the original image $\mathbf I$ into $N_s$ superpixels using the simple linear iterative clustering algorithm~\cite{slic}. Then, we randomly fill a fixed number $S_s$ of the superpixels with random colors to generate a distorted image $\mathbf{I_{surf}}$. In addition, we produce a binary mask $\mathbf{M_{surf}}$ that localizes the randomly filled superpixels. An example of this is shown in Fig.~\ref{FIG2:model} (a). 

\begin{table*}[!t]
\begin{center}
\begin{adjustbox}{width=1.4\columnwidth,center}
\begin{tabular}{c|cccccccc|c}
\hline
Categories & GeoT~\cite{geotrans} & GAN~\cite{ganomaly} & US~\cite{student}   & ITAE~\cite{itae} & SPADE~\cite{spade} & PSVDD~\cite{patchsvdd} & RIAD~\cite{riad} & DivA~\cite{hou2021divide} & TSDN \\ \hline\hline
Texture    & 58.5     & 76.5     & 91.5 &  82.2  & -  & 94.5 & \textcolor{blue}{95.1}  & 91.0 &  \textcolor{red}{98.2}    \\
Object     & 71.6     & 75.4     & 85.8 &  84.8 & - & \textcolor{red}{90.8} & 89.9  & 88.8 &   \textcolor{blue}{90.1}   \\
All     & 67.2     & 76.2     & 87.7 &  83.9  &85.5 &\textcolor{blue}{92.1} & 91.7  & 89.5 &   \textcolor{red}{92.8}   \\ \hline
\end{tabular}
\end{adjustbox}
\vspace{-0.3cm}
\caption{Comparison of image-level anomaly detection accuracy on MVTec\cite{mvtec} (AUC $\%$). The table shows the averages of the texture and object categories. The top two results in each category are shown in \textcolor{red}{red} and \textcolor{blue}{blue}, respectively.}
\label{tab2}
\vspace{-0.9cm}
\end{center}
\end{table*}

\vspace{-0.4cm}
\subsection{Two Stream Decoder Network Architecture}
\vspace{-0.25cm}
We propose a TSDN to effectively discriminate and eliminate the abnormal features. From the deepest layer of our encoder, we apply $1 \times 1$ convolution to reduce the channels. We call this feature $\mathbf{F_{1/32}} \in \mathbb{R}^{C \times  \frac{ W } { 32 }\times \frac{ H } { 32 }}$, where $C$ indicates the number of channels of $ \mathbf{F_{1/32}}$, and $W$ and $H$ are the width and height of $\mathbf{I_{surf}}$, respectively. $\mathbf{F_{1/32}}$ is then fed to $DCD_A$, which is composed of deconvolution blocks and has skip connections between the encoder layers to supplement the information lost during feature extraction. $DCD_A$ learns the abnormal features within $\mathbf{F_{1/32}}$ to generate $\mathbf{M'_{surf}}$, the predicted binary anomaly mask.

Next, $\mathbf{F_{1/32}}$ and $\mathbf{M'_{surf}}$ are fed into the proposed FNE, which is to be further presented in Sec.~\ref{section3:FNE}. The FNE estimates the similarity between $\mathbf{M'_{surf}}$ and each channel of $\mathbf{F_{1/32}}$. The output of FNE is $\mathbf{W'} \in \mathbb{R}^{C}$, a vector of the similarity distance scores. The more similar, the smaller the score, and vice versa. 
We multiply $\mathbf{W'}$ to $\mathbf{F_{1/32}}$ and obtain a new feature $\mathbf{F_{nml}} = \mathbf{W'} \times \mathbf{F_{1/32}}$.


 Because the channels of $\mathbf{F_{1/32}}$ that are highly similar to the predicted anomaly mask $\mathbf{M'_{surf}}$ have low values in their corresponding $\mathbf{W'}$, these channels are depressed in $\mathbf{F_{nml}}$. Therefore, $\mathbf{F_{nml}}$ contains only the normal features, directly solving the problem of conveying excessive features.

$DCD_N$ is a decoder that generates $\mathbf{R_{surf}}$, a reconstruction of the anomaly-free image $\mathbf I$. Because the abnormal features are eliminated in $\mathbf{F_{nml}}$, $DCD_N$ is encouraged to replace the lost features with the adjacent features to generate an anomaly-free output $\mathbf{R_{surf}}$. 
This also makes $DCD_N$ robust in learning normal features. Consequently, $\mathbf{R_{surf}}$ will have precise details in the normal areas, whereas it will have lower quality in the abnormal areas, thus succeeding in discriminating between the normal and abnormal areas.

\vspace{-0.3cm}
\subsection{Feature Normality Estimator}
\vspace{-0.2cm}
\label{section3:FNE}
We propose the FNE to distinguish and eliminate abnormal features within $\mathbf{F_{1/32}}$. 
Fig.~\ref{FIG2:model} (b) illustrates the FNE process. First, we generate the ground truth $\mathbf {W_{gt}}$ to train the FNE. $\mathbf{F_{1/32}}$ and $\mathbf{M'_{surf}}$ are resized to match each other. $\mathbf{F_{1/32}}$ is upscaled to the size of $\mathbb{R}^{C \times { W }\times { H }}$, and $\mathbf{M'_{surf}}$ in expanded to have $C$ channels. Then, we compute the channel-wise distance between $\mathbf{F_{1/32}}$ and $\mathbf{M'_{surf}}$ using the structural similarity index (SSIM)~\cite{ssim} as the metric. The equation is:

\vspace{-0.6cm}
\begin{equation}
\begin{split}
		\label{ssim}
		D(\mathbf{F_{1/32}}, \mathbf{M'_{surf}}) = SSIM(\mathbf{F_{1/32}}, \mathbf{M'_{surf}})
\end{split}
\end{equation}
\vspace{-0.6cm}

\noindent Then, we normalize $D(\mathbf{F_{1/32}}, \mathbf{M'_{surf}})$ to the range $[0, 1]$ to obtain $\mathbf{W_{gt}}$. Consequently, the channels of $\mathbf{F_{1/32}}$ that have a structure similar to $\mathbf{M'_{surf}}$ are given low values, and vice versa. Thereby, we obtain a distance score vector $\mathbf {W_{gt}} \in \mathbb{R}^{C}$. The length of $\mathbf{W_{gt}}$ is the same as the number of channels of $\mathbf{F_{1/32}}$, indicating that each value of $\mathbf{W_{gt}}$ has a one-to-one correspondence to the channels of $\mathbf{F_{1/32}}$.

To train the FNE, we apply convolution to $\mathbf{F_{1/32}}$ and flatten it. The flattened output is fed into a three-layer MLP that estimates the normality score of each channel of $\mathbf{F_{1/32}}$. The first two layers consist of a fully connected layer and rectified linear unit activation. The last layer has a fully connected layer, and we apply a sigmoid function to generate a vector $\mathbf {W'}$. The FNE learns to estimate the normality of each channel by generating $\mathbf {W'}$. The training is optimized to minimize the cross-entropy loss $L_{FNE}$ between $\mathbf{W_{gt}}$ and $\mathbf{W'}$.

\vspace{-0.3cm}
\subsection{Optimization}
\vspace{-0.2cm}
We optimize our model with five objective functions: $L_{r}$, $L_{s}$, $L_{g}$, $L_{m}$, and $L_{FNE}$. $L_r$ is defined by the pixel-wise $L_2$ distance between $\mathbf I$ and $\mathbf{R_{surf}}$ as $L_{r} \left (\mathbf{I}, \mathbf{R_{surf}} \right) = |\mathbf{I}, \mathbf{R_{surf}}|$.
We also use the SSIM~\cite{ssim} distance $L_{s}$ and gradient magnitude similarity (GMS)~\cite{gms} $L_{g}$ between $\mathbf I$ and $\mathbf{R_{surf}}$. These are used in addition to $L_{r}$ to promote perceptual similarity. 
Furthermore, we compute the $L_2$ distance between $\mathbf{M_{surf}}$ and $\mathbf{M'_{surf}}$ expressed as: $L_{m} \left (\mathbf{M_{surf}}, \mathbf{M'_{surf}} \right) = |\mathbf{M_{surf}}, \mathbf{M'_{surf}}|$. As described in Sec.~\ref{section3:FNE}, $L_{FNE}$ is the cross-entropy loss between $\mathbf{W_{gt}}$ and $\mathbf{W'}$ (Eq.~\ref{Lfne}).

\vspace{-0.3cm}
\begin{equation}
	\label{Lfne}
	L_{FNE} \left (\mathbf{W_{gt}}, \mathbf{W'} \right) = CEL(\mathbf{W_{gt}}, \mathbf{W'})
\vspace{-0.2cm}
\end{equation}

\noindent The final loss function is obtained by combining the five functions which are multiplied by $\lambda_r$, $\lambda_s$, $\lambda_g$, $\lambda_m$, and $\lambda_f$, the weights controlling the contribution. 

\vspace{-0.6cm}
\begin{equation}
	\label{loss}
	L_{total} =  \lambda_r L_{r} +\lambda_s L_{s} + \lambda_g L_{g} + \lambda_m L_{m} + \lambda_f L_{FNE},
\vspace{-0.2cm}
\end{equation}

\vspace{-0.4cm}
\subsection{Anomaly Prediction}
\vspace{-0.2cm}
\noindent {\bf Pixel-level.} To detect pixel-level anomalies, we adopt the difference between $\mathbf I$ and $\mathbf{R_{surf}}$ as the score map. We compute the $L_2$ distance and apply Gaussian blurring to obtain a smoothed version of the difference map $S_{map}$. Such smoothed map is more robust toward small and poorly reconstructed anomalous regions. $S_{map}$ is expressed as:

\vspace{-0.3cm}
\begin{equation}
	\label{scoremap}
	S_{map} \left (\mathbf{I}, \mathbf{R_{surf}} \right) = G(|\mathbf{I}, \mathbf{R_{surf}}|),
\vspace{-0.15cm}
\end{equation}

\noindent where G denotes the Gaussian blur function. Lastly, we normalize $S_{map}$ to the range $[0, 1]$ to obtain the score map $S_{final}$.

\noindent {\bf Image-level.} To detect anomalies at the image-level, we leverage the maximum value of $S_{map}$ of each image as the abnormality score and normalize the scores to the range $[0, 1]$. Where the higher the score, the more potentially abnormal. 

\vspace{-0.3cm}
\section{Experiments}
\vspace{-0.3cm}
\subsection{Implementation Details}
\vspace{-0.2cm}
We implemented our experiments using a single NVIDIA RTX A6000. We used the Adam optimizer with a learning rate of $7e^{-5}$. We adopted EfficientNet-b5~\cite{tan2019efficientnet}, pretrained on ImageNet, as our backbone. We loaded frames in the RGB and resized them to $352 \times 352$ pixels.
Furthermore, we empirically set $N_s$ and $S_s$ to $400$ and $50$, respectively. The weights for the loss function in Eq.~\ref{loss} were set to $\lambda_r=1, \lambda_s=5e^{-1}, \lambda_g=1, \lambda_m=1,$ and  $\lambda_f=5e^{-5}$ to balance the loss values for stable training.

\begin{table}[!t]
    \begin{center}
    \begin{adjustbox}{width=0.9\columnwidth,center}
    \begin{tabular}{m{0.3cm}|c|cccccc|c}
        \toprule  
        \multicolumn{2}{c}{Categories} & $AE_{S}$~\cite{mvtec} & VAE~\cite{vevae} & SMAI~\cite{smai} & PNET~\cite{pnet}&  RIAD~\cite{riad} & GP~\cite{gp}  & TSDN\\ 
        \midrule
        \midrule
        \multirow{6}{*}{\rot{Textures}} &
        Carpet &	87.0	&	78.0	&	87.0 	&57.0 &\textcolor{red}{96.3}	& \textcolor{blue}{96.0} & 93.9	\\
        & Grid &	94.0	&	73.0	&	96.0 	&98.0 &\textcolor{blue}{98.8}	& 78.0 & \textcolor{red}{98.9}	\\
        & Leather &	78.0	&	95.0	&	51.0 	&89.0 &\textcolor{red}{99.4}	& 90.0 & \textcolor{blue}{99.1}	\\
        & Tile &	59.0	&	80.0	&	60.0	&\textcolor{blue}{97.0}  & 89.1 & 80.0 & \textcolor{red}{97.5}	\\
        & Wood &	73.0	&	77.0	&	62.0	&\textcolor{red}{98.0} &85.8	& 81.0 & \textcolor{blue}{93.7}	\\
        \cline{2-9}
        & $avg_{tex}$ & 78.2	&	80.6	&	71.2 	& 87.8	& \textcolor{blue}{93.8} & 85.0 & \textcolor{red}{96.5}	\\
        \hline
        \multirow{11}{*}{\rot{Objects}}
        & Bottle &	93.0	&	87.0 &	91.0 & \textcolor{red}{99.0} &\textcolor{blue}{98.4}	& 93.0 & 86.2	\\
        & Cable &	82.0	&	\textcolor{red}{90.0}	&	82.0 & 70.0 &84.2 & \textcolor{blue}{94.0} & 86.6	\\
        & Capsule &	94.0	&	74.0	&	81.0 & 84.0	&\textcolor{blue}{92.8} & 90.0 & \textcolor{red}{94.7}	\\
        & Hazelnut &	97.0	&	\textcolor{red}{98.0}	&	96.0	& \textcolor{blue}{97.0}	&96.1 & 84.0 & \textcolor{red}{98.0}	\\
        & Metal nut &	89.0	&	\textcolor{blue}{94.0}	&	90.0 	& 79.0	&92.5 & 91.0 & \textcolor{blue}{92.6}	\\
        & Pill &	91.0	&	83.0	&	93.0 	& 91.0	&\textcolor{red}{95.7} & 93.0 & \textcolor{blue}{93.4}	\\
        & Screw &	96.0	&	97.0	&	94.0 	& \textcolor{red}{100.0} &\textcolor{blue}{98.8} & 96.0 & 97.3	\\
        & Toothbrush &	92.0	&94.0	& 	96.0 	& \textcolor{red}{99.0} &98.9 	& 96.0 & 91.0	\\
        & Transistor &	90.0	&	93.0	&	82.0 	& 82.0	&87.7 & \textcolor{red}{100.0} & 73.3	\\
        & Zipper &	88.0	&	78.0	&	74.0 	&\textcolor{blue}{97.8} & \textcolor{red}{99.0}	& \textcolor{red}{99.0} & 93.7	\\
        \cline{2-9}						
        & $avg_{obj}$ & 87.0	&	86.1	&	87.9 & 90.0	&\textcolor{red}{94.4} & \textcolor{blue}{93.6} & 90.8	\\
         \bottomrule
    \end{tabular}
    \end{adjustbox}
    \vspace{-0.3cm}
    \caption{Comparison of pixel-wise anomaly localization accuracy on MVTec~\cite{mvtec} (AUC $\%$).}
    \label{tab1}
    \vspace{-0.9cm}
    \end{center}
\end{table}

\noindent {\bf Evaluation metrics.} For a fair comparison with the existing works, we adopted the image-level and pixel-level area under the curve (AUC) of the receiver operating characteristics curve. These metrics are used in most studies~\cite{mvtec, gp, vevae, smai, pnet}.

\noindent {\bf Dataset.} We evaluated our model with the MVTecAD~\cite{mvtec}, a widely used benchmark designed to estimate the abnormality localization performance in industrial quality control. This dataset is composed of five texture categories and $10$ object categories. This dataset is used in most of the recent studies~\cite{vevae, pnet, mvtec, gp, patchsvdd, riad}. Though some works~\cite{vevae, gp, geotrans} have also evaluated on datasets such as Cifar10~\cite{krizhevsky2009learning} or MNIST~\cite{deng2012mnist}, we do not consider using these because these are not specifically designed for our target task, as their abnormal images come from a completely different category.

\vspace{-0.4cm}
\subsection{Quantitative Results}
\vspace{-0.3cm}
\noindent {\bf Image-level anomaly detection.} Table~\ref{tab2} quantitatively compares the image-level detection accuracy against eight state-of-the-art methods~\cite{geotrans, ganomaly, student, itae, spade, patchsvdd, hou2021divide, riad}. We show the averages of the texture and object categories as well as all categories combined. 
Our TSDN outperforms all others on the texture categories, surpassing the other models.\

\noindent {\bf Pixel-level anomaly segmentation.} Table~\ref{tab1} shows our anomaly segmentation performance on the MVTec dataset~\cite{mvtec}. We find that our TSDN surpasses or performs competitively in all $15$ categories. Particularly for the texture categories, TSDN is more accurate than the counterpart methods, ranking first or second in four out of five categories. 

\begin{table}[!t]
\begin{center}
\begin{adjustbox}{width=1.0\linewidth,center}
\begin{tabular}{l|ccc|clccccc}
\hline
  & $DCD_N$ & \begin{tabular}[c]{@{}c@{}}$DCD_A$\\ +FNE\end{tabular} & Skip & Carpet & Grid & Hazelnut & Metal nut & Screw & Toothbrush & Zipper \\ \hline
  \hline
A & \ding{51}                                                 &                                                    &                                                           & 87.3   & 98.0 & 97.6     & 79.6      & 97.0  & 91.3       & 90.1   \\
B & \ding{51}                                                 &                                                    & \ding{51}                                                         & 70.7   & 90.1 & 92.0     & 70.5      & 94.9  & 87.6       & 84.4   \\
C &             \ding{51}                                      & \ding{51}                                                  &                                                           &    92.0    &  98.2    &    91.1      &      90.4     &    93.0   &   93.0         &    91.2    \\
D &            \ding{51}                                       & \ding{51}                                                  & \ding{51}                                                         & \bf{93.9}   & \bf{98.5} & \bf{98.0}     & \bf{92.6}      & \bf{97.6}  &     \bf{94.8}       & \bf{93.7}   \\ \hline
\end{tabular}
\end{adjustbox}
\vspace{-0.3cm}
\caption{The impact of our architecture with respect to the skip connections. We show the pixel-level AUC ($\%$). }
\label{tab3:ablation}
\vspace{-0.8cm}
\end{center}
\end{table}

\begin{table}[!t]
\begin{center}
\begin{adjustbox}{width=1.0\linewidth,center}
    \begin{tabular}{l|ccccc|cccccc}
    \hline
                           &$DCD_N$ & Skip & SURF & $DCD_A$ & FNE & Screw & Hazelnut & Zipper & Metal nut & Wood & Pill \\ \hline\hline
    A & \ding{51}                                                 &                                                           &      &      &     & 97.0  & 97.6     & 90.1   & 79.6                                                & 89.9 & 92.0 \\
    B & \ding{51}                                                 &                                                           & \ding{51}    &      &     & 97.2  & 97.7     & 90.8   & 80.6                                                & 90.3 & 92.5 \\
    C & \ding{51}                                                 & \ding{51}                                                         &      &      &     & 94.9  & 92.0     & 84.4   & 70.5                                                & 90.3 & 87.8 \\
    D & \ding{51}                                                 & \ding{51}                                                         & \ding{51}    &      &     & 97.0  & 97.7     & 88.8   & 80.6                                                & 90.6 & 91.5 \\
    E & \ding{51}                                                 & \ding{51}                                                         & \ding{51}    & \ding{51}    &     & 97.3  & 97.7     & 88.8   & 91.1                                                & 90.8 & 91.4 \\
    F & \ding{51}                                                 & \ding{51}                                                         & \ding{51}    & \ding{51}    & \ding{51}   & \bf{97.6}  & \bf{98.0}     & \bf{93.7}   & \bf{92.6}                                                & \bf{93.7} & \bf{93.4} \\ \hline
\end{tabular}
\end{adjustbox}
\vspace{-0.3cm}
\caption{The impact of our proposed modules. We show the pixel-wise accuracy (AUC $\%$) of five experiments.}
\label{tab4}
\end{center}
\vspace{-0.9cm}
\end{table}

\vspace{-0.4cm}
\subsection{Ablation Studies}
\vspace{-0.2cm}
\noindent {\bf Using skip connections.} 
Table~\ref{tab3:ablation} shows our model’s ability to use skip connections. 
We show its performance in the following four experiments: (A) a one-stream basic reconstructive AE, (B) an AE with skip connections, (C) our model without skip connections, and (D) our model with skip connections. Table~\ref{tab3:ablation} (B) shows that the AE with skip connections fails to benefit from the conveyed features of the skip connections because the performance deteriorates compared to the AE without them, as shown in Table~\ref{tab3:ablation} (A). 
The indiscriminate conveying of features involves those of the abnormal regions, leading to the abnormal appearance left in the reconstructed output. In contrast, our TSDN (Table~\ref{tab3:ablation} (D))  uses the skip connections effectively by distinguishing the useful normal features and eliminating the abnormal features. 

\noindent {\bf Impact of our SURF and FNE.} We also conducted ablation studies regarding the proposed SURF and FNE, as presented in Table~\ref{tab4}. 
First, it is worth noting the impact of the FNE. Table~\ref{tab4} (E) shows the performance of our model without the FNE. For this experiment, we replaced the FNE by simply multiplying $\mathbf{M'_{surf}}$ by $\mathbf{F_{1/32}}$, intended to remove the learning of discrimination of the normal feature channels. Observations from this model reveal that the proposed FNE remarkably improves the pixel-wise performance by approximately $2.36$ percent points on average. 
Moreover, even without the FNE, our two-stream decoder design with $DCD_A$ and $DCD_N$, shown in Table~\ref{tab4} (E), still shows improvement over the one-stream AE, shown in Table~\ref{tab4} (D). 
Additionally, SURF is beneficial to the reconstruction-based anomaly detection, as the performance in Tables~\ref{tab4} (B) and (D) improves over that in Tables~\ref{tab4} (A) and (C), respectively. 
This reveals that SURF helps optimize the model to the distinguishable abnormal images by training it in the same setting as the testing.

	\begin{figure}[!t]
		\begin{center}
			\includegraphics[width=0.9\linewidth]{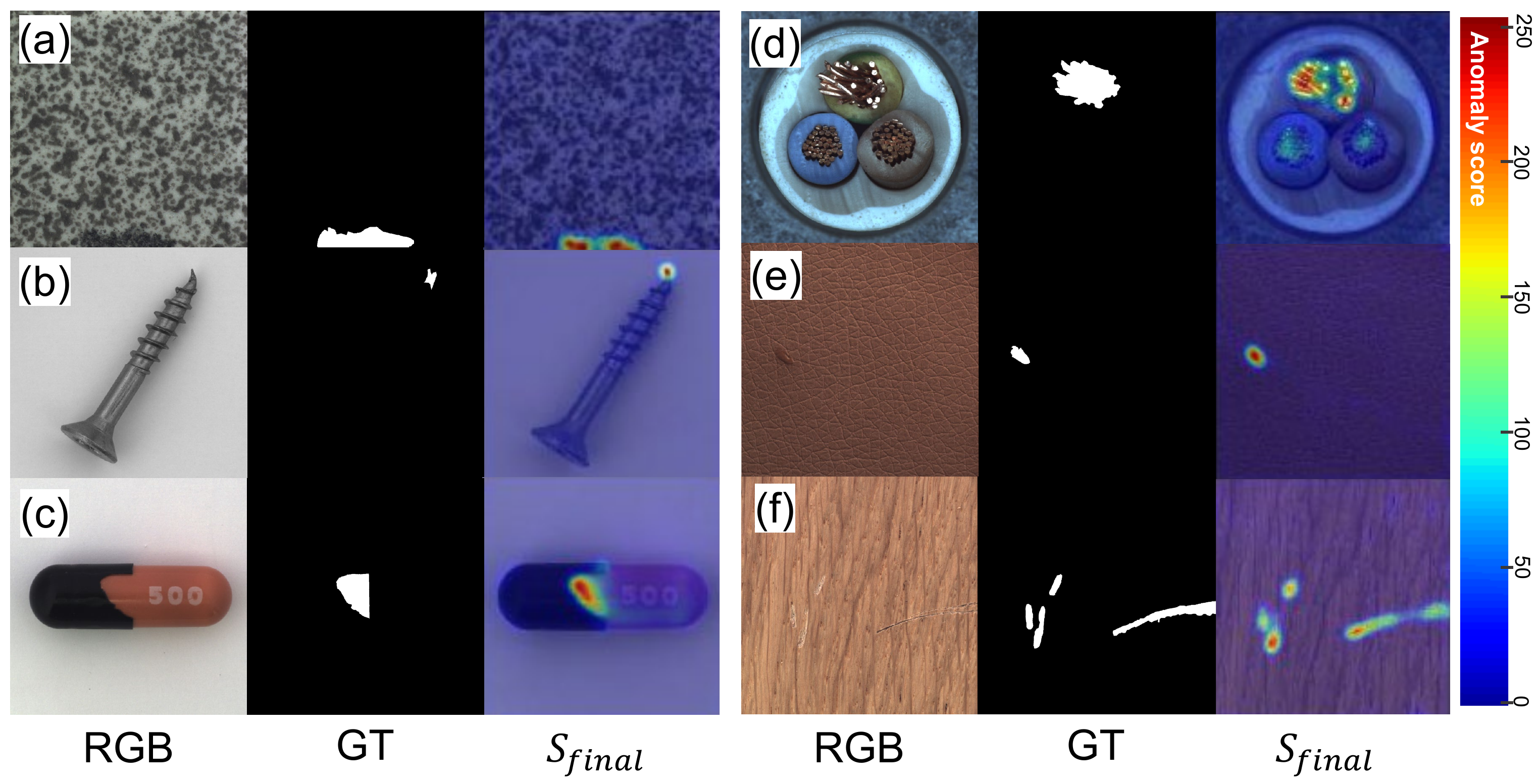}
		\end{center}
		\vspace{-0.6cm}
		\caption{Qualitative results on the MVTec dataset~\cite{mvtec}.}
		\label{FIG7:result}
	\vspace{-0.4cm}
	\end{figure}

	\begin{figure}[!t]
		\begin{center}
			\includegraphics[width=0.6\linewidth]{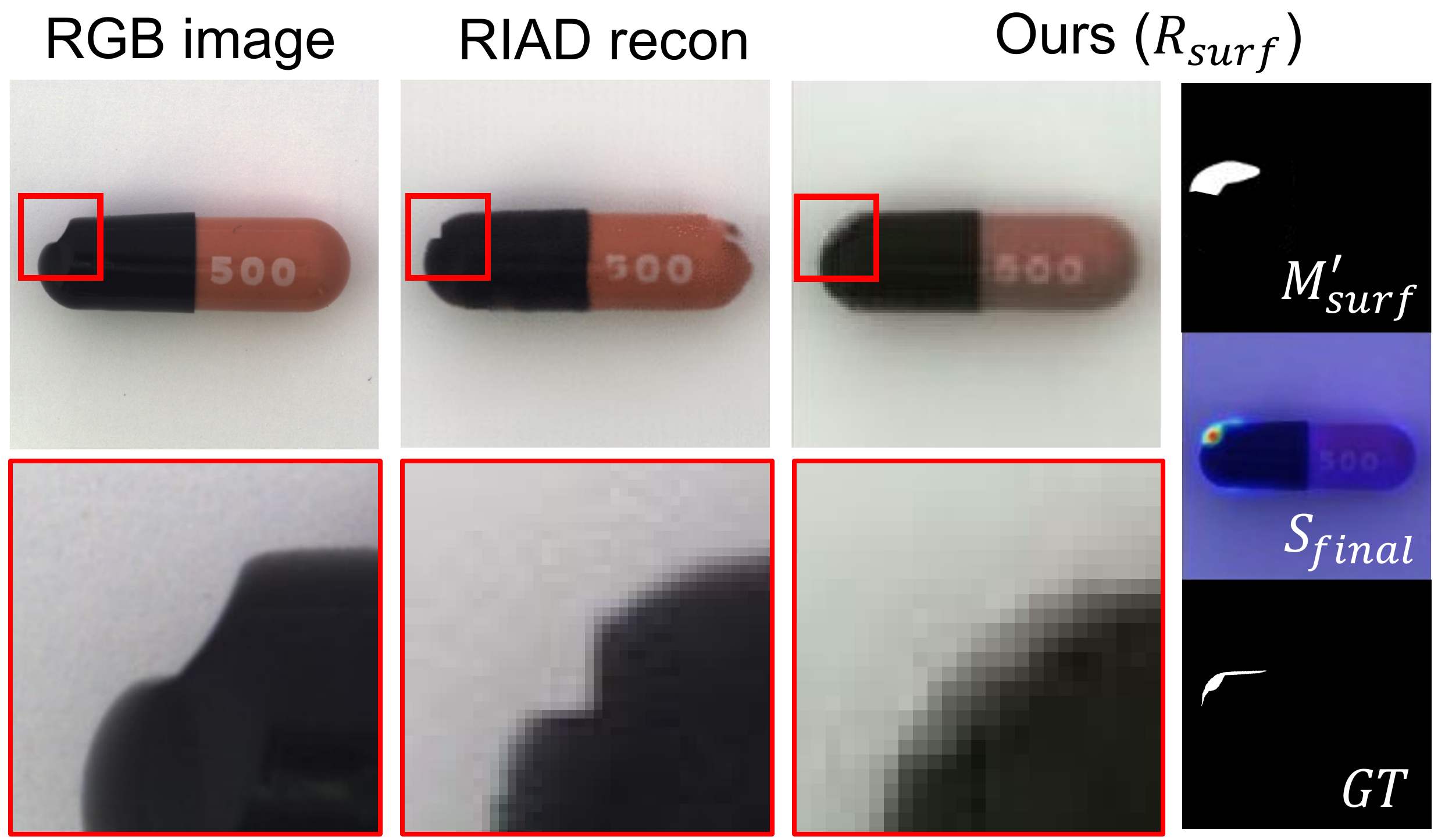}
		\end{center}
		\vspace{-0.5cm}
		\caption{Visualization of $R_{surf}$, $M'_{surf}$, and the score map $S_{final}$. We compared $R_{surf}$ with the output of RIAD~\cite{riad}.
        }
        \vspace{-0.6cm}
		\label{FIG8:result}
	\end{figure}

\vspace{-0.3cm}
\subsection{Qualitative Results}
\vspace{-0.2cm}
Fig.~\ref{FIG7:result} presents the score maps $S_{final}$ of our model. The results in Figs.~\ref{FIG7:result} (b) and (e) show that our model can segment tiny targets. 
Moreover, our model is robust to complex anomalies that are difficult to distinguish from the background, as shown in Figs.~\ref{FIG7:result} (a), (d), and (f). Also, (b) and (c) show that our model is also applicable to structural anomalies. 

Furthermore, Fig.~\ref{FIG8:result} demonstrates $R_{surf}$ and $M'_{surf}$ generated from $DCD_N$ and $DCD_A$, respectively. We also present the output of RIAD~\cite{riad}, which is also a transformation-based method, but that uses a single-stream AE. 
This figure validates the effect of the two-stream decoder design. 
The $R_{surf}$ shows that our model reconstructed the squeezed pill to a more normal appearance. 
In contrast, the RIAD~\cite{riad} reconstruction clearly retains the squeezed appearance. 

\vspace{-0.5cm}
\section{Conclusion}
\vspace{-0.3cm}
In this paper, we aimed to segment anomalies by alleviating the powerful generalization problem of previous methods due to indiscriminate conveying of features. Specifically, we proposed TSDN, which captures both normal and abnormal features. Furthermore, the proposed FNE effectively eliminates the abnormal features, helping to prevent the reconstruction of the abnormal regions. Extensive experiments clearly demonstrated the effectiveness of our method.

\newpage
\begingroup
\setstretch{0.9}
\bibliographystyle{IEEEbib}
\bibliography{refs}
\endgroup

\end{document}